\documentclass{article}
\usepackage{spconf,amsmath,graphicx,amssymb,amsfonts}
\usepackage[caption=false,font=normalsize,labelfont=sf,textfont=sf]{subfig}
\usepackage[hyphens]{url}

\title{High-Capacity Complex Convolutional Neural Networks for I/Q Modulation Classification}

\name{Jakob Krzyston$^{\star}$ \thanks{Email: jakobk@gatech.edu, Code: https://github.com/JakobKrzyston/} \qquad Rajib Bhattacharjea$^{\dagger}$ \qquad Andrew Stark$^{\star}$}
  
\address{$^{\star}$ School of Electrical and Computer Engineering, Georgia Institute of Technology\\
      $^{\dagger}$ DeepSig Inc.\\}

\begin{document}
%
\maketitle
\begin{abstract}
I/Q modulation classification is a unique pattern recognition problem as the data for each class varies in quality, quantified by signal to noise ratio (SNR), and has structure in the complex-plane. Previous work shows treating these samples as complex-valued signals and computing complex-valued convolutions within deep learning frameworks significantly increases the performance over comparable shallow CNN architectures. In this work, we claim state of the art performance by enabling high-capacity architectures containing residual and/or dense connections to compute complex-valued convolutions, with peak classification accuracy of 92.4\% on a benchmark classification problem, the RadioML 2016.10a dataset. We show statistically significant improvements in all networks with complex convolutions for I/Q modulation classification. Complexity and inference speed analyses show models with complex convolutions substantially outperform architectures with a comparable number of parameters and comparable speed by over 10\% in each case.
\end{abstract}
\begin{keywords}
Complex convolution, modulation classification, deep learning, I/Q modulation, telecommunications
\end{keywords}

\section{Background}
O'Shea et al. \cite{o2016convolutional} first introduced deep learning approaches to I/Q modulation pattern classification, showing great accuracy improvements over traditional statistical methods. In \cite{o2018over}, a thorough exercise in various methods for radio signal classification was conducted and showed deep learning architectures such as ResNets significantly outperform advanced statistical machine learning methods such as gradient boosted trees with hand crafted high-order statistical features. In \cite{ramjee2019fast}, Ramjee et al. furthered this work by trying many different deep learning approaches to classify modulated radio signals and investigated as how to best reduce the training time of the approaches. Deep learning architectures investigated included a  DenseNet, and Convolutional Long Short-term Deep Neural Network (CLDNN), ResNet and the LSTM architectures. Their results showed the ResNet outperformed all architecture over all SNRs tested. Ramjee et al. demonstrated that, in general, more sophisticated architectures outperform those demonstrated by O'Shea et al. since they are capable of mapping new non-linear relationships in addition to having many more parameters to optimize over.

This problem space is generally aimed at improving the modulation classification task developed in [1]. However, direct comparison is difficult because: (a) inconsistency in reporting train/test split on a dataset~\cite{west2017deep,luo2019radio}, (b) inconsistency in how the train/test sets are formed, whether the data is randomly shuffled or not,~\cite{xu2020spatiotemporal}, and (c) use of a simpler dataset~\cite{lin2020hybrid,wu_sun_wei_zhao} which has many more samples per class and fewer classes. To our best knowledge, few publications make any effort toward repeatability or reproducibility. 

In this study we test our methods against approaches that have performed competitively, if not state-of-the-art, in the literature. We demonstrate our methods substantially outperform these approaches; peak classification accuracy of 92.4\% in complex models compared to a peak of 83.6\%, outperforming parameter-matched models by over 10\%, and outperforming models with equivalent inference speed by over 10\%.

\section{Complex Convolutions for Real-Valued Inputs}
\label{our_way}
Convolutional neural networks do not actually compute convolutions, but rather, cross-correlations. In applications such as telecommunications, computing a convolution in the complex domain is important to filter a complex data stream (known as I/Q data) and extract features for classification. Treating a two dimensional array consisting of real and imaginary components as a two dimensional array consisting of real values does not allow for the network to learn from the joint correlations that inherently exist in I/Q data~\cite{krzyston2020complex, krzyston_icpr}.

Recent work~\cite{krzyston2020complex} showed how real-valued deep learning frameworks can compute complex-valued convolutions using standard deep-learning convolutional followed by a linear combination. Implementing complex convolutions into the architecture from~\cite{o2016convolutional} drastically improved the classification accuracy. Further work~\cite{krzyston_icpr}, showed complex convolutions are able to better learn feature representations in noisy environments and are more effective than naively adding more parameters to a network.  


Recently, there has been another approach developed to enable deep learning paradigms to compute complex convolutions~\cite{chakraborty2019surreal}. This work takes a geometric approach to understanding the relationship between the real and imaginary components by defining the convolution as a weighted Fr\'{e}chet mean on a Lie group. This new form of convolution necessitated the development of a new activation function, $G$-transport. When comparing to the work of \cite{o2016convolutional}, this method of complex convolutions slightly underperformed~\cite{o2016convolutional} while producing a model that was 70\% the size. 

For comparison, our method to compute complex convolutions, originally demonstrated in~\cite{krzyston2020complex}, used 1.0038 times the number of parameters as the network in~\cite{o2016convolutional}. Although our complex convolutions use twice as many parameters as traditional convolutions, this slight increase stems from the majority of the parameters being in the dense layers. However, our method outperformed~\cite{o2016convolutional} with statistical significance over five trials~\cite{krzyston_icpr}. Further, in~\cite{krzyston_icpr} the activation maximizations were qualitatively analyzed and it can be seen that the features learned with complex convolutions better captures the relationship between I and Q components than traditional CNNs.

\section{High-Capacity CNN Architectures with Complex Convolutions}
Krzyston et al. have only tested complex convolutions by linear combination~\cite{krzyston2020complex, krzyston_icpr} on a low-capacity CNN described in~\cite{o2016convolutional}. The CNN in~\cite{o2016convolutional} has two convolutional layers and two dense layers. The network in \cite{krzyston2020complex}, called Krzyston 2020, only differs in the first convolutional layer being complex.

Following the work of AlexNet in 2012~\cite{NIPS2012_4824}, there were many follow up works trying to evaluate the relationship between the depth of a network and improvements in performance \cite{simonyan2014very, szegedy2015going}. However, gradients would get smaller as networks get deeper preventing deep architectures from learning, better known as the vanishing gradient problem.

The vanishing gradient problem was ultimately addressed by Residual Networks~\cite{he2016deep}, better known as ResNets. ResNets immediately became the state of the art across computer vision, and remain the leading architecture to use for image classification and pattern recognition problems. Following the work of Residual Networks, was the development of Densely Connected Networks~\cite{huang2017densely}, known as DenseNets.

In this work, we integrate complex convolutions into state of the art CNN paradigms, Residual Networks, Densely Connected Networks, as well as a combination of the two, which we call a Dense ResNet, inspired by~\cite{liublind}.

\subsection{Residual Networks}
\begin{figure}
\centering
\includegraphics[width=2 in]{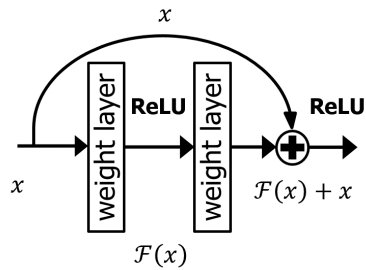}
\caption{Residual connection, originally described in~\cite{he2016deep}.}
\label{fig:res}
\end{figure}

Residual connections~\cite{he2016deep} in neural networks were proposed to address the problem of vanishing gradients and enable networks to be deeper than originally though possible. By adding the identity to itself, it enabled features to propagate further through the network, suggesting more robust representation learning in addition to proving to be easier to optimize. Figure~\ref{fig:res} shows how a residual connection is formed.

Utilizing the PyTorch ResNet GitHub repo~\cite{pytorch_res}, we develop two smaller ResNets, ResNet-18 and ResNet-34. We developed smaller ResNets to keep inference speeds low. We developed two different sizes to test if more layers leads to a meaningful improvement in classification performance. The number indicates the number of computational layers, counted in the same manner of more well-known ResNets. We developed the complex convolution variants, named ResNet-18 C and ResNet-34 C respectively, where all convolutional layers compute complex convolutions.

\subsection{Densely Connected Networks}
\begin{figure}
\centering
\includegraphics[width=3.39 in]{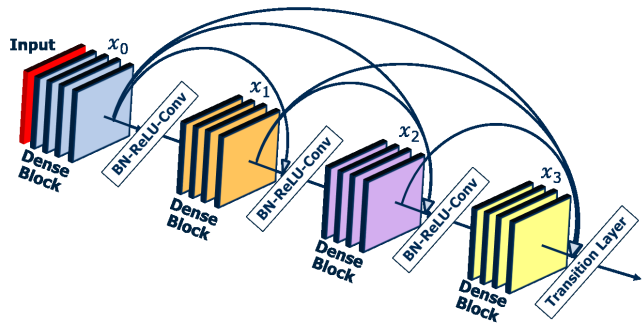}
\caption{DenseNet architecture, originally described in~\cite{huang2017densely}, with only four dense blocks.}
\label{fig:dense}
\end{figure}

Dense connections~\cite{huang2017densely} in neural networks were proposed as an alternative to residual connections for the vanishing gradients problem while adding far fewer parameters to optimize than ResNets. In dense connections, each 'block', comprised of convolutional layers in series, is connected to every other block and the feature maps from one layer are concatenated to the input of the next. Figure~\ref{fig:dense} shows a DenseNet comprised of four densely connected dense blocks.

Utilizing the PyTorch DenseNet GitHub repo~\cite{pytorch_dense}, we develop two smaller DenseNets and their complex convolution variants, DenseNet-57 (C) and DenseNet-73 (C).

\subsection{Dense ResNets}
Inspired by the work done in~\cite{liublind}, we developed an architecture that utilizes both dense and residual connections, Dense ResNets. Dense ResNets leverage both means of addressing the vanishing gradient problem and feature propagation. Following~\cite{liublind}, the Dense ResNets have 6 blocks, each block is comprised of four of the residual connections seen in Figure~\ref{fig:res} (totaling eight convolutional layers per block), and are followed by two dense layers. Additionally the kernel size decreases every other block, from 7, to 5, to 3. 

With the same motivation for two smaller ResNets and DenseNets, we developed two sizes of the Dense ResNet, Dense ResNet-35 and Dense ResNet-68, the difference being the number of blocks. Dense ResNet-68 has six blocks and the Dense ResNet-35 has three blocks, each with descending kernel sizes. The number indicates the total number of computational layers in the architecture and is counted in the same fashion as a ResNet or DenseNet architecture. Complex convolutional variants were created, denoted as Dense ResNet-35 C and Dense ResNet-68 C respectively.

\section{Experimental Design}
These architectures were trained and tested on the RadioML 2016.10a open source dataset used in \cite{o2016convolutional}. This standard baseline dataset for I/Q modulation classification consists of 11 modulations (8 digital and 3 analog) at SNR levels from -20 to 18 dB with 2 dB steps. Additionally the dataset includes variation in the following properties: center frequency offset, sample clock rate, sample clock offset, and initial phase. There 1,000 samples of all modulation schemes at all SNR values. The shape of each sample is 2 x 128, representing 128 samples in time and 2 channels, I and Q \cite{o2016convolutional}.

\begin{figure}
\centering
\includegraphics[width=3.39 in]{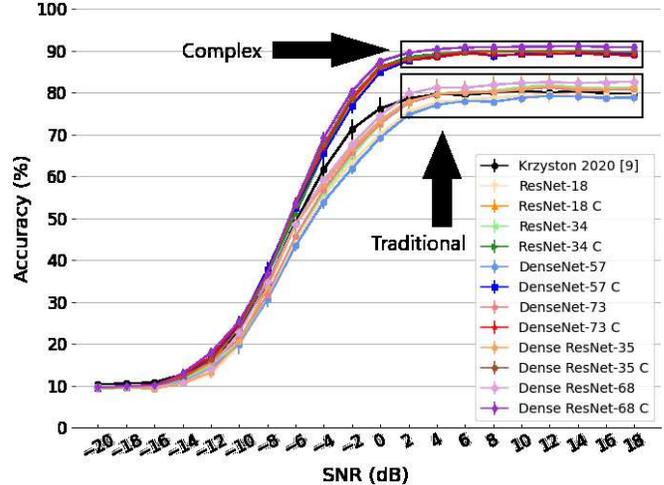}
\caption{Average classification accuracy as a function of SNR with standard deviation bars. The boxes enclose the accuracies of the complex and traditional convolutional high-capacity architectures respectively.}
\label{fig:class_all}
\end{figure}

\begin{figure*}
\centering
\includegraphics[width=4.3 in]{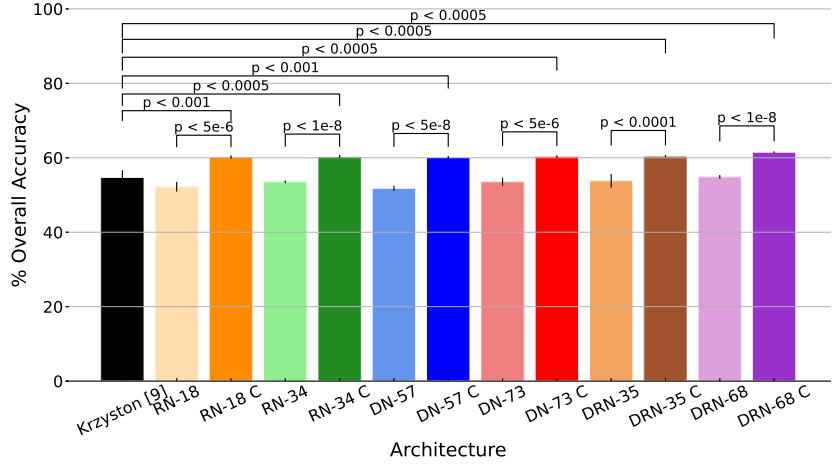}
\caption{Average overall accuracies, with standard deviation bars, for all models tested. The acronyms are as follows: Krzyston is from [9], RN = ResNet, DN = DenseNet, DRN = Dense ResNet, and C = utilized complex convolutions. The unpaired student t-test was used to compute p-values determining statistical significance of one models performance versus another.}
\label{fig:summary}
\end{figure*}

The data was shuffled across both the modulation formats and SNR levels then split 50/50 into train/test sets. Each architecture performed the classification task five times, reshuffling to obtain new train/test sets each time. Inference speed was quantified as the duration for a trained network to make a prediction with the test samples and trained model loaded onto the GPU. All trials were performed on a single NVIDIA GeForce GTX 1080 Ti GPU.

\section{Results}

In Figure~\ref{fig:class_all}, the averaged classification accuracy with standard deviation bars is plotted as a function of SNR. Boxed and shown by the darker colored plots, all of the complex variations of all high-capacity architectures outperform their traditional convolution counterparts, especially as SNR increases above -6dB. For all SNR greater than -2dB these complex variation increase by at least 12\%. Further, the addition of skip connections enabled the complex networks to perform much better than the Krzyston 2020 network. Krzyston 2020 outperforms all of the higher capacity networks utilizing traditional convolutions, when tested on samples below 2dB SNR. The Dense ResNet-68 C achieved state of the art performance with an average overall accuracy of 61.5\%, peaking at 92.4\% at 14 dB SNR. It achieves over 80\% accuracy for SNRs greater than -2dB, only requiring 128 digital I/Q samples.

Figure~\ref{fig:summary} compares the classification accuracies of the complex convolutional networks, over five trials, versus the traditional variants as well as the Krzyston 2020 network. The performances of all the high-capacity complex architectures outperformed their traditional counterparts and the Krzyston 2020 architecture with statistical significance. The unpaired student t-test was used to compute p-values.

\begin{figure}
\centering
\includegraphics[width=3.39 in]{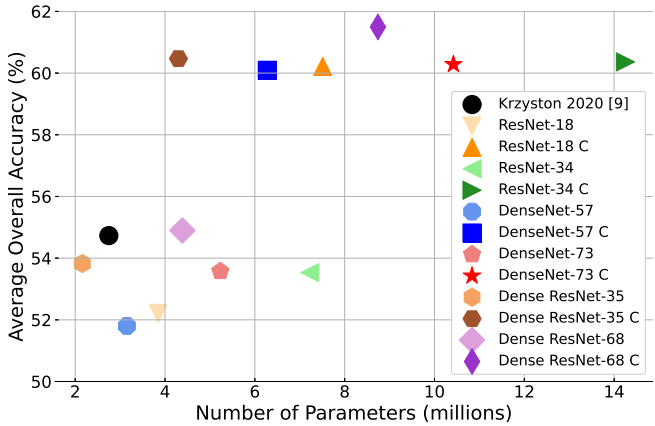}
\caption{Average classification accuracy plotted against the number of parameters in the architecture, in millions.}
\label{fig:param_v_acc}
\end{figure}

Figure~\ref{fig:param_v_acc} shows the relationship between average overall accuracy and the number of trainable parameters for each model. High-capacity architectures nearly doubled in complexity when enabled to compute complex convolutions, due to the large number of convolutional layers in each architecture and the complex convolutions leveraging a 2 x $m$ kernel size. However, when comparing models with nearly the same number of parameters, the ability to perform complex convolutions substantially improves the performance. For example, the Dense ResNet-68 and Dense ResNet-35 C models have nearly the same number of parameters as well as the ResNet-34 and ResNet-18 C architectures being nearly parameter-matched, yet in both cases the architecture with the ability to compute complex convolutions outperforms the other, on average, by 10.15\% and 12.49\% respectively. Further, Figure~\ref{fig:param_v_acc} shows there was no substantial performance benefit from adding more layers/parameters to the models.

\begin{figure}
\centering
\includegraphics[width=3.39 in]{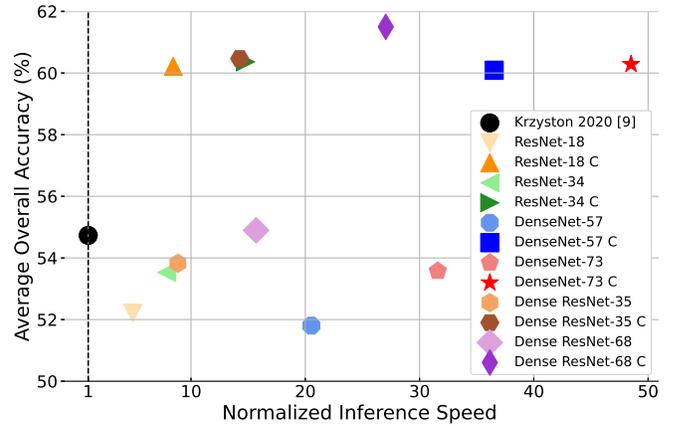}
\caption{Average classification accuracy plotted against average inference speeds, normalized by [9] (0.398 $\mu$s/sample).}
\label{fig:speed_v_acc}
\end{figure}

Figure~\ref{fig:speed_v_acc} shows the relationship between average overall accuracy and average normalized inference speed. The average inference speeds for each network was normalized to the average inference speed of the Krzyston 2020 network, which was 0.398$\mu$s/test sample. Complex models tend to take longer to compute inferences. However, a complex network of equal inference speed greatly outperforms a traditional architecture of similar speed. For example, the Dense ResNet-68 and Dense ResNet-35 C models compute inferences at the same speed but the Dense ResNet-35 C outperforms by 10.15\%. 

\section{Conclusion}
In this work we examined the modulation classification performance of high-capacity architectures when enabled to compute complex convolutions. Combining complex convolutions and various types of skip connections enables state of the art performance on I/Q modulation classification. An architectures ability to compute complex convolutions yields over 10\% higher accuracy than simply using a large architecture with the same number of parameters or an architecture that infers at the same speed.

Future work includes speeding up the performance of these higher-capacity, complex convolutional networks for real-world applications via quantization/pruning. Further, recent articles~\cite{spooner_2020_bpsk,spooner_2020_more} detailed fundamental issues with the RML2016a and RML2016b datasets, which are commonly used in the field. The impacts of these issues on classification performance and generalizability of the trained models is an area of future investigation.

\vfill
\pagebreak

\Urlmuskip=0mu plus 1mu\relax
\bibliographystyle{IEEETran}
\bibliography{bib_JK}

\end{document}